\documentclass[sigconf,natbib=true]{acmart}
\usepackage[T1]{fontenc}
\usepackage{graphicx}
\usepackage{xcolor}  
\usepackage{amsmath}
\usepackage{algorithm}
\usepackage{algpseudocode}
\usepackage{hyperref}   
\usepackage{hyperxmp}   

\AtBeginDocument{%
  }
\begin{document}

\title{STRIVE: A Think \& Improve Approach with Iterative Refinement for Enhancing Question Quality Estimation}


\author{Aniket Deroy}
\orcid{0000−0001−7190−5040}
\affiliation{
 \institution{IIT Kharagpur}
\state{West Bengal}
\country{India}
}
\email{roydanik18@kgpian.iitkgp.ac.in}
\authornote{Corresponding Author}

\author{Subhankar Maity}
\orcid{0009−0001−1358−9534}
\affiliation{%
 \institution{IIT Kharagpur}
      \state{West Bengal}
  \country{India}
}
\email{subhankar.ai@kgpian.iitkgp.ac.in}

\begin{abstract}
Automatically assessing question quality is crucial for educators as it saves time, ensures consistency, and provides immediate feedback for refining teaching materials. We propose a novel methodology called \textbf{STRIVE} (\textbf{\underline{S}}tructured \textbf{\underline{T}}hinking and \textbf{\underline{R}}efinement with multi-LLMs for \textbf{\underline{I}}mproving \textbf{\underline{V}}erified Question \textbf{\underline{E}}stimation) using a series of Large Language Models (LLMs) for automatic question evaluation. This approach aims to improve the accuracy and depth of question quality assessment, ultimately supporting diverse learners and enhancing educational practices.
The method estimates question quality in an automated manner by generating multiple evaluations based on the strengths and weaknesses of the provided question and then choosing the best solution generated by the LLM. Then the process is improved by iterative review and response with another LLM until the evaluation metric values converge. This sophisticated method of evaluating question quality improves the estimation of question quality by automating the task of question quality evaluation. Correlation scores show that using this proposed method helps to improve correlation with human judgments compared to the baseline method. Error analysis shows that metrics like relevance and appropriateness improve significantly relative to human judgments by using STRIVE.
\end{abstract}

\ccsdesc[500]{Computing methodologies~Natural language generation}
\ccsdesc[500]{Applied computing~Education}

\keywords{Education, Large Language Models (LLMs), Question Generation, Automated Question Assessment, Feedback, GPT }


\maketitle

\section{Introduction}
Automatically assessing question quality~\cite{deutsch2021towards,das2021automatic,ni2022deepqr} is significant for educators for several reasons. First, it saves time, allowing educators to focus on teaching rather than question formulation~\cite{NEURIPS2023_f323d594}. Second, automated systems provide a standardized measure of question quality~\cite{wang2021results,min2021neurips}, reducing subjectivity and bias in evaluations. In addition, educators can receive immediate feedback on the effectiveness of their created questions~\cite{ni2022deepqr,chang2022webqa}, allowing the continuous refinement of the teaching materials.
Information retrieval via questions~\cite{gaur2022iseeq,zamani2020generating,bonifacio2022inpars} empowers learners to engage with knowledge more interactively and effectively. Integrating question-driven information retrieval into education can foster curiosity, critical thinking, and tailored learning experiences, creating a dynamic and accessible educational ecosystem.
Here, the task is to automatically assess metrics such as grammaticality (i.e., Gram), relevance (i.e., Rel), appropriateness (i.e., App), complexity (i.e., Com), and novelty (i.e., Nov) in questions generated from educational materials \cite{maity2023harnessing}. Moreover, analyzing question quality can help educators identify gaps in curriculum coverage or areas where students may struggle, leading to more targeted instruction. 
We know that human evaluations of questions are costly. So we ask the question, \textit{Whether a novel method comprising of LLMs can replace human evaluations in terms of assessing educational question quality?}

In this work, our contributions are as follows:
(i) We propose a novel method called STRIVE that leverages the capabilities of LLMs to generate multiple evaluations based on each question's inherent strengths and weaknesses within the context of the question generation task. STRIVE identifies the most effective question assessment by assessing these evaluations, thus enhancing the quality of educational assessments. Furthermore, the method incorporates an iterative review process, utilizing two \textit{Think and Improve} modules (TM) comprising LLMs to refine the evaluations until the results align between the two \textit{\textit{Think and Improve}} modules. (ii) We compare our baseline method with our proposed approach, STRIVE, and show that our approach achieves scores closer to the human baseline. The correlation scores~\cite{sedgwick2012pearson,benesty2008importance} show that our proposed approach STRIVE aligns better with human experts. The error analysis~\cite{lauer2013question,wang2009computer,cai2020errors,araki2016generating} shows relevance and appropriateness is considerably improved using our proposed approach STRIVE.



\section{Dataset}
We used 1000 <Context, Question> pairs from the EduProbe~\cite{maity2023harnessing} dataset for the purpose of our study. The dataset comprises of subjects like History, Geography, Environmental Studies, and Economics from 6th standard to 12th standard according to NCERT curriculum. We pick up 1000 contexts from the SciQ dataset~\cite{welbl2017crowdsourcing} and create 1000 <Context, Question> pairs from Chemistry, Biology, Physics, Earth Sciences, etc. We hired an educator to create question from every context. We use these <Context, Question> pairs for our study. The context and open-ended questions are used for the purpose of evaluations via the LLMs. To show the wider applicability of our work, we test our proposed approach STRIVE on two different datasets namely EduProbe and SciQ covering a wide range of subjects like History, Geography, Environmental Studies, Economics, Chemistry, Biology, Physics, Earth Sciences, etc.

\section{Methodology}
\subsection{Baseline Approach}
\begin{figure}[h!]
  \centering
  \includegraphics[width=0.65\linewidth]{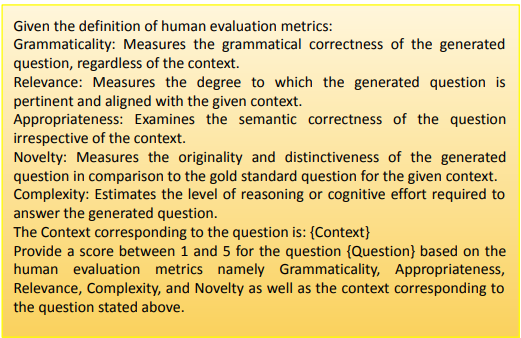}
  \caption{Prompt for evaluating metrics as per baseline method.} \label{fig1}
\end{figure}

The methodology for the baseline approach is shown in Figure~\ref{fig4}. The corresponding prompt for baseline approach is shown in Figure~\ref{fig1}. The algorithm for baseline approach is shown in Algorithm \ref{al0}.


\begin{algorithm}
\caption{Automated Question Quality Evaluation Using Baseline Approach}
\begin{algorithmic}[1]
\Require Human Evaluation Metric definitions
\Require Question and its context
\Ensure Human evaluation metric scores for the question

\State{Provide Human evaluation metric definitions, context,question as input to LLM}

\Return Final Human evaluation metric scores

\end{algorithmic}
\label{al0}
\end{algorithm}





\subsection{STRIVE Approach}
We first discuss the \textit{Think and Improve} modules (i.e., ${TM}_1$ and ${TM}_2$) present in the STRIVE approach\footnote{The phrases '\textbf{STRIVE Approach}' and '\textbf{Feedback-based Approach}' are used interchangeably in this section and subsequent sections.}. 
There are two steps in the \textit{\textit{Think and Improve}} module: (a) In the first step, we generate multiple strengths and flaws for the question. We generate 10 different <strength, weakness> pairs for every question at different temperature values. 
(b) In the second step, we ask the LLM to choose the best set of strength from the available set of all strengths generated by the LLM in the previous step. We follow a similar methodology for selecting the weakness set. 
In addition to selecting the best set of strengths and weaknesses, we also calculate the metric scores corresponding to the evaluation metrics.

\begin{figure}[h!]
  \centering
  \includegraphics[width=0.65\linewidth]{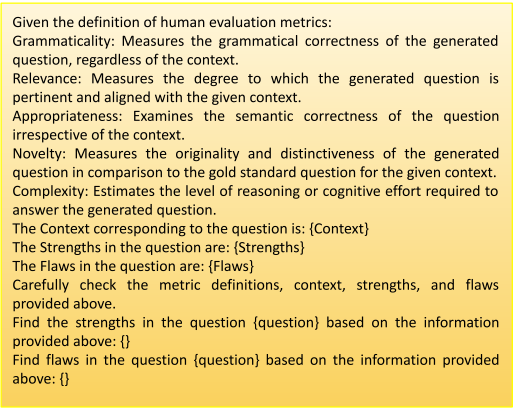}
  \caption{Prompt for generating multiple strength and weakness pairs in \textit{Think and Improve} module.} \label{fig2}
\end{figure}

\begin{figure}[h!]
  \centering
  \includegraphics[width=0.65\linewidth]{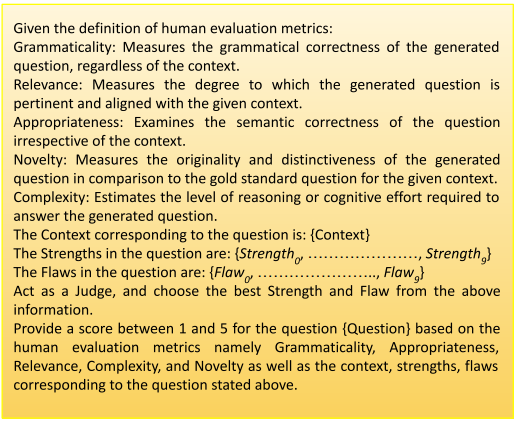}
  \caption{Prompt for choosing the best strength and weakness pairs in the \textit{Think and Improve} module and generating metric scores.} \label{fig3}
\end{figure}

\begin{figure}[h!]
  \centering
  \includegraphics[width=0.65\linewidth]{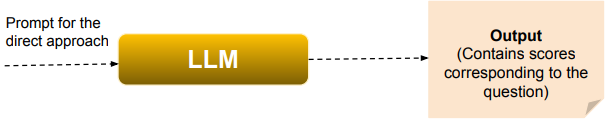}
  \caption{Overview of the Baseline Approach.} \label{fig4}
\end{figure}

Figure~\ref{fig6} represents the overview for \textit{Think and Improve} module for generating multiple strength and weakness pairs and then choosing the best strength and weakness pair from the set.
Figure~\ref{fig2} shows the $Prompt_{gen}$ used in Figure~\ref{fig6} to generate strength and weakness pairs from context and question.
Figure~\ref{fig3} represents $Prompt_{judge}$,  also used in Figure~\ref{fig6}. The prompt in Figure \ref{fig3} helps select the best strength and weakness pair from all available pairs and provides the evaluation metric scores.

\begin{figure}[h!]
  \centering
  \includegraphics[width=0.90\linewidth]{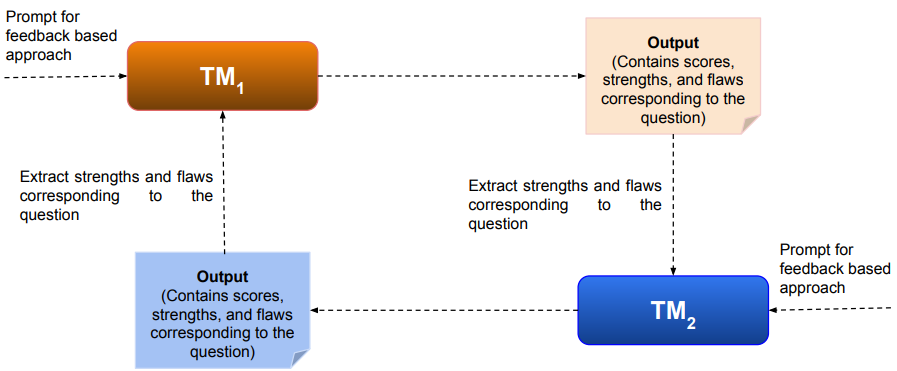}
  \caption{Overview of the STRIVE Approach.} \label{fig5}
\end{figure}

\begin{figure}[h!]
  \centering
  \includegraphics[width=0.80\linewidth]{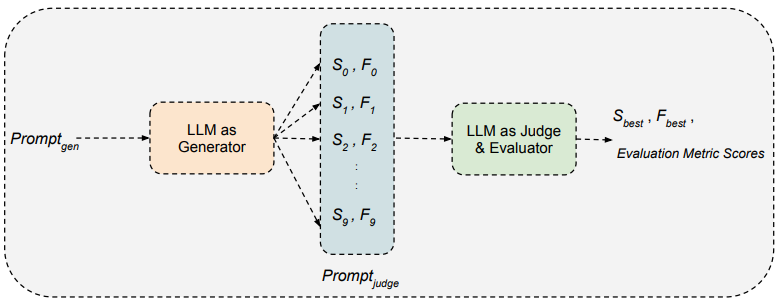}
  \caption{Overview of the Think \& Improve Module (i.e., TM).} \label{fig6}
\end{figure}

\begin{algorithm}
\label{al_1}
\caption{Automated Question Quality Evaluation Using STRIVE Approach}
\begin{algorithmic}[1]
\Require Human Evaluation Metric definitions
\Require Initial strengths set $S_0$ and flaws set $F_0$, both initialized as empty sets
\Require Question and its context
\Ensure Converged scores for the question

\State $S_0 \gets \{\}$
\State $F_0 \gets \{\}$

\State $S_0, F_0 \gets \text{Identify 10 sets of strengths and flaws}$

\State $S_0, F_0 \gets \text{Use LLM as a judge to choose the best strength and flaw}$

\State Compute initial scores for human evaluation metrics with the given question and context

\While{convergence criteria are not met}
    
    \State Provide $S_0, F_0$ as feedback, along with metric definitions, question, and context, to ${TM}_2$

    \State $S_{1} \gets \text{Generate 10 variations of Strengths} $
    
    \State $F_{1} \gets \text{Generate 10 variations of Flaws} $

    \Comment{Select best strength and flaw}
    
    \State $S_{1} \gets \text{Select best strength from } S_{1}$
    
    \State $F_{1} \gets \text{Select best flaw from } F_{1}$

Generate scores for metrics by ${TM}_2$\\

    \State Provide $S_1$, $F_1$ as feedback, along with metric definitions, question, and context, to ${TM}_1$
    
    
    \Comment{Generate variations of strengths and flaws from ${TM}_1$}
    \State $S_{2} \gets \text{Generate 10 variations of Strength} $
    \State $F_{2} \gets \text{Generate 10 variations of Flaws}$

    \Comment{Select best strength and flaw from ${TM}_1$}
    \State ${S_2} \gets \text{Select best strength from } S_{2}$
    \State ${F_2} \gets \text{Select best flaw from } F_{2}$

\State Ask ${TM}_1$ to generate scores for human evaluation metrics

\State ${S_0} \gets  S_{2}$
   \State ${F_0} \gets  F_{2}$


    \Comment{Check for convergence}
    \If{scores from ${TM}_1$ and ${TM}_2$ are identical for two consecutive iterations}
        \State Terminate the loop
    \EndIf
\EndWhile

\Return final scores and strengths/flaws
\end{algorithmic}
\label{al_1}
\end{algorithm}
The algorithm \ref{al_1} outlines an iterative process for evaluating human evaluation metrics using two \textit{Think and Improve} modules consisting of LLMs, referred to as ${TM}_1$ and ${TM}_2$ (See Figure \ref{fig5}). It begins by requiring definitions for the human evaluation metrics and initializing two empty sets: $S_0$ for strengths and $F_0$ for flaws. The algorithm takes a specific question and its context as input. Initially, it computes initial strengths $S_0$ and flaws $F_0$ consisting of 10 pairs of strengths and flaws via LLMs and then again uses the LLM as a judge to decide the most suitable strength and flaw from all the possible strengths and flaws. Then the human metric scores corresponding to the question and context is calculated. ${TM}_1$ takes into consideration this entire evaluation process. During each iteration, the identified best strength and flaw are provided as feedback from ${TM}_1$ to ${TM}_2$, along with the metric definitions, question, and context. ${TM}_2$ then generates ten variations of the strengths and flaws. From these variations, the LLM selects the best strength and the best flaw. ${TM}_2$ also calculates the scores for the evaluation metrics. ${TM}_2$ sends the best strength and flaw to ${TM}_1$. The process iteratively continues until the evaluation metric scores generated by ${TM}_1$ and ${TM}_2$ are exactly same for all the evaluation metrics for two consecutive iterations. Figure~\ref{fig5} shows the figure for the iterative approach for the STRIVE algorithm via two \textit{Think and Improve} modules. The iterative process continues until the evaluation metric scores converge between the two modules for two subsequent runs.

\section{Results}

\begin{table}[h]
    \centering
    \caption{Model performance metrics for the EduProbe dataset. Values in \textcolor{blue}{blue} represent the highest value for each metric and approach.}
    \scalebox{0.60}{ 
        \begin{tabular}{lccccc}
            \toprule
            \textbf{Model} & \textbf{Gram} & \textbf{App} & \textbf{Rel} & \textbf{Nov} & \textbf{Com} \\
            \midrule
            Human Baseline & 4.86 & 4.87 & 4.53 & 3.75 & 3.23 \\
            \midrule
            \multicolumn{6}{l}{\textbf{EduProbe (Baseline Approach)}} \\ \midrule
            GPT-4 & \textcolor{blue}{4.71} & \textcolor{blue}{4.67} & \textcolor{blue}{4.18} & \textcolor{blue}{3.56} & \textcolor{blue}{3.12} \\
            Gemini-Pro & 4.58 & 4.61 & 4.12 & 3.48 & 3.03 \\
            Llama3 & 4.48 & 4.34 & 3.89 & 3.34 & 3.07 \\
            \midrule
            \multicolumn{6}{l}{\textbf{EduProbe (Feedback-based Approach)}} \\ \midrule
            GPT-4 & \textcolor{blue}{4.82} & \textcolor{blue}{4.81} & \textcolor{blue}{4.37} & \textcolor{blue}{3.65} & \textcolor{blue}{3.20} \\
            Gemini-Pro & 4.62 & 4.67 & 4.21 & 3.63 & 3.08 \\
            Llama3 & 4.57 & 4.52 & 4.12 & 3.40 & 3.10 \\
            \bottomrule
        \end{tabular}
    }
    \label{tab:Eduprobe}
\end{table}

\begin{table}[h]
    \centering
    \caption{Model performance metrics for the SciQ dataset. Values in \textcolor{blue}{blue} represent the highest value for each metric and approach.}
    \scalebox{0.7}{
        \begin{tabular}{lccccc}
            \toprule
            \textbf{Model} & \textbf{Gram} & \textbf{App} & \textbf{Rel} & \textbf{Nov} & \textbf{Com} \\
            \midrule
            Human Baseline & 4.84 & 4.74 & 4.25 & 3.80 & 4.09 \\
            \midrule
            \multicolumn{6}{l}{\textbf{SciQ (Baseline Approach)}} \\ \midrule
            GPT-4 & \textcolor{blue}{4.24} & \textcolor{blue}{4.52} & \textcolor{blue}{4.14} & \textcolor{blue}{3.54} & \textcolor{blue}{3.53} \\
            Gemini-Pro & 4.22 & 4.44 & 4.12 & 3.50 & 3.24 \\
            Llama3 & 4.03 & 4.21 & 3.85 & 3.45 & 3.19 \\
            \midrule
            \multicolumn{6}{l}{\textbf{SciQ (Feedback-based Approach)}} \\ \midrule
            GPT-4 & \textcolor{blue}{4.60} & \textcolor{blue}{4.71} & \textcolor{blue}{4.23} & \textcolor{blue}{3.65} & \textcolor{blue}{3.81} \\
            Gemini-Pro & 4.42 & 4.51 & 4.22 & 3.60 & 3.65 \\
            Llama3 & 4.34 & 4.42 & 4.03 & 3.54 & 3.71 \\
            \bottomrule
        \end{tabular}
    }
    \label{tab:SciQ}
\end{table}

Table~\ref{tab:Eduprobe} shows model performance metrics for the EduProbe dataset. Table~\ref{tab:SciQ} shows model performance metrics for the SciQ dataset.
\begin{table}[h]
    \centering
    \caption{Pearson correlation coefficient between GPT-4 generated scores and human experts. Values in \textcolor{blue}{blue} represent the highest value for each metric and dataset.}
    \scalebox{0.8}{ 
        \begin{tabular}{lccccc}
            \toprule
            \textbf{Model} & \textbf{Gram} & \textbf{App} & \textbf{Rel} & \textbf{Nov} & \textbf{Com} \\ 
            \midrule
            \multicolumn{6}{l}{\textbf{EduProbe}} \\ \midrule
            GPT-4 (Baseline Approach) & 0.38 & 0.41 & 0.26 & 0.28 & 0.29 \\
            GPT-4 (Feedback-based Approach) & \textcolor{blue}{0.44} & \textcolor{blue}{0.62} & \textcolor{blue}{0.42} & \textcolor{blue}{0.63} & \textcolor{blue}{0.56} \\
            \midrule
            \multicolumn{6}{l}{\textbf{SciQ}} \\ \midrule
            GPT-4 (Baseline Approach) & 0.33 & 0.41 & 0.28 & 0.29 & 0.28 \\
            GPT-4 (Feedback-based Approach) & \textcolor{blue}{0.61} & \textcolor{blue}{0.56} & \textcolor{blue}{0.47} & \textcolor{blue}{0.51} & \textcolor{blue}{0.45} \\
            \bottomrule
        \end{tabular}
    }
    \label{tab:gpt4_performance}
\end{table}
Human baseline scores are the highest across all metrics, indicating that human evaluators provide the most grammatically accurate (i.e., Gram), appropriate (i.e., App), relevant (i.e., Rel), novel (i.e., Nov), and complex (i.e., Com) responses. GPT-4 shows competitive performance, but generally falls short of the human baseline, particularly in novelty for the baseline approach. Gemini-Pro scores lower than GPT-4. Llama3 scores lowest in all metrics within this approach, indicating less effective performance compared to the other models for the baseline approach.

For the feedback-based approach, GPT-4 scores improve compared to the baseline approach, especially in relevance and appropriateness, demonstrating the effectiveness of feedback in improving model performance.
Gemini-Pro also shows improved scores in the STRIVE approach, particularly in relevance and appropriateness, although it still does not match GPT-4.
Llama3 shows slight improvements, but still trails behind the other LLMs. Our proposed approach, STRIVE seems to enhance performance, especially for GPT-4. For the SciQ dataset, the human baseline performs relatively well across all metrics, with the highest score in grammaticality and the lowest in novelty. For SciQ, GPT-4 shows moderate performance with a high score in grammaticality but lower scores in novelty and complexity for the baseline approach.
Gemini-Pro has similar trends, with a strong appropriateness score but lower novelty and complexity.
Llama3 consistently scores the lowest among the baseline models, particularly in novelty and complexity. For the feedback-based approach, the performance generally improves across all metrics compared to the baseline approach.
GPT-4 achieves higher scores in all human evaluation metrics than in the baseline approach.
Gemini-Pro also improves, especially in grammaticality and relevance. Llama3 performance improves compared to the baseline approach.

Table~\ref{tab:gpt4_performance} shows the Pearson correlation coefficient scores between GPT-4 and human experts. We hired 16 human experts who are experienced educators from the \href{https://www.upwork.com/}{UpWork} platform and distributed all the questions among the 16 human experts so that they could provide a human evaluation score for every question. For the EduProbe dataset, GPT-4 (Baseline Approach) shows relatively low correlation values, with the highest being for appropriateness and the lowest being for relevance. 
GPT-4 (Feedback-based Approach) shows improved correlations in all metrics.
For SciQ, we observed a similar trend with the STRIVE approach (i.e., feedback-based approach) outperforming the baseline approach for all the correlation values across all the metrics.
In summary, these correlation values highlight the effectiveness of our proposed feedback-based approach in improving performance in different metrics.

\section{Error Analysis}
We selected 100 questions from the SciQ dataset and 100 questions from the EduProbe dataset. We checked what percentage of values predicted by human experts which match exactly with the LLM-generated values for the different metrics.
\begin{table}[h]
    \centering
    \caption{Percentage of score matches between LLM and human experts. Values in \textcolor{blue}{blue} represent the highest value for each metric.}
    \scalebox{0.8}{
    \begin{tabular}{@{}lccccc@{}}
        \toprule
        \textbf{Model} & \textbf{Gram} & \textbf{App} & \textbf{Rel} & \textbf{Nov} & \textbf{Com} \\
        \midrule
        GPT-4 (Baseline Approach) & 60.5 & 45.0 & 40.5 & 50.5 & 48.0 \\
        GPT-4 (Feedback-based Approach) & \textcolor{blue}{74.5} & \textcolor{blue}{71.0} & \textcolor{blue}{63.0} & \textcolor{blue}{61.0} & \textcolor{blue}{68.5} \\ \midrule
        Gemini Pro (Baseline Approach) & 55.0 & 40.5 & 35.0 & 42.5 & 44.0 \\
        Gemini Pro (Feedback-based Approach) & 70.0 & 69.0 & 61.5 & 55.5 & 62.0 \\ \midrule
        Llama3 (Baseline Approach) & 50.5 & 35.5 & 32.0 & 38.5 & 36.5 \\
        Llama3 (Feedback-based Approach) & 68.5 & 60.5 & 54.0 & 48.5 & 50.0 \\
        \bottomrule
    \end{tabular}
    }
    \label{tab:error}
\end{table}
Table~\ref{tab:error} shows the percentage of score matches between LLM and human experts. We observe that several exact matches in scores between the LLM-provided scores and human experts improve in the case of the feedback-based approach as compared to the baseline approach for three LLMs namely, GPT-4, Gemini pro, and Llama3. The highest match between human experts and LLM is observed in grammaticality and appropriateness. The lowest match is observed for novelty between LLM and human experts.
We observe that the percentage matches between GPT-4 (Feedback-based Approach) and human experts improve significantly for appropriateness and relevance.

\section{Related Work}
Evaluating questions~\cite{wang2024revisiting} is a time-consuming and critical task given the reasoning that evaluating the questions~\cite{zamani2020generating,claveau2021neural,reddy2022entity,yang2019hybrid} automatically is challenging. Various aspects of question quality~\cite{madans2011question} like grammaticality, relevance, novelty, complexity, and appropriateness need to be measured, which is difficult to do manually. Automated metrics often fail to capture nuanced aspects~\cite{gorgun2024exploring,mulla2023automatic} of what makes a question "good" (e.g., relevance, clarity, or cognitive challenge). Automated metrics like BLEU and ROUGE rely on n-gram overlaps and ignore semantic quality or rephrasing.
They do not account for the appropriateness or answerability of the question~\cite{ji2022qascore,horbatiuk2023test} within the given context. Metrics~\cite{moreno2024evaluating,mulla2023automatic} like BLEU~\cite{papineni2002bleu} and ROUGE~\cite{lin2004rouge} focus on surface-level similarity and cannot deeply understand, whether the question aligns with the context provided. So, there is not much work on automated evaluation of question generation using human evaluation-like metrics. In this work, we aim to fill this gap.


\section{Conclusion}
In conclusion, the ability to automatically assess question quality plays a vital role in modern education by streamlining the evaluation process. The novel STRIVE approach, using LLMs represents a significant advancement in this area. By generating multiple evaluations of a given question and iteratively refining these assessments, STRIVE provides a more accurate, nuanced understanding of the question quality. This approach not only supports educators in identifying areas for improvement, but also fosters better alignment with human judgment in terms of relevance, appropriateness, and overall effectiveness. The strong correlation between STRIVE-generated evaluations and human assessments underscores the potential of this method to enhance educational practices. Additionally, error analysis shows strong improvements in relevance and appropriateness in our proposed feedback-based approach (i.e., STRIVE) compared to the baseline approach.
\bibliographystyle{ACM-Reference-Format}
\bibliography{sample-base}









\end{document}